# Identifying Optimal Sequential Decisions

**A. Philip Dawid**
Statistical Laboratory,
Centre for Mathematical Sciences,
Cambridge, UK
apd@statslab.cam.ac.uk

**Vanessa Didelez**
Department of Mathematics
University of Bristol, UK
vanessa.didelez@bristol.ac.uk

## Abstract

We consider conditions that allow us to find an optimal strategy for sequential decisions from a given data situation. For the case where all interventions are unconditional (atomic), identifiability has been discussed by Pearl & Robins (1995). We argue here that an optimal strategy must be conditional, i.e. take the information available at each decision point into account. We show that the identification of an optimal sequential decision strategy is more restrictive, in the sense that conditional interventions might not always be identified when atomic interventions are. We further demonstrate that a simple graphical criterion for the identifiability of an optimal strategy can be given.

## 1 INTRODUCTION

Consider the case of a chronically ill patient who regularly sees their doctor in order to adjust their treatment to the individual development of their disease. For example patients who receive anticoagulation treatment are subject to regular blood testing; depending on the outcome of such a test, and possibly on earlier blood tests, the dosage of the anticoagulant might be modified. An optimal strategy in this context is a rule that stipulates, for each point in time where an intervention is carried out, a dosage to be given so as to optimise the treatment outcome, e.g. to keep the coagulation measure stable within certain bounds. It is intuitively obvious that to achieve optimality such a rule will typically have to be a function of the individual patient's history, e.g. for a patient whose blood coagulates too fast the dose has to be increased while for a patient whose blood coagulates too slowly it has to be decreased. Decision strategies where each intervention is allowed to depend on the individual history are called *conditional*, *dynamic* or *adaptive treatment* strategies. A particular feature of such strategies is that for a given patient it is not known what exact dose they will receive at future interventions as this will depend on their future coagulation test results which are subject to random variation and influenced by many factors other than treatment, such as diet and lifestyle.

Here we consider the question of what kind of data situation will allow us to construct, i.e. identify, an optimal decision strategy. This is particularly relevant when data is obtained from observational studies, but also informs us about the design of experimental studies. Essentially we need to be able to identify all conditional strategies over which we want to optimise. Identifiability of unconditional sequential strategies (atomic interventions) has been considered by Pearl & Robins (1995), where a graphical criterion is given to read identifiability off a causal diagram. Dawid & Didelez (2005) generalise their graphical criterion to the case where some or all of the interventions are allowed to depend on some or all of the observable previous information. Our main result here is that if *all* interventions are allowed to depend on *all* of the observable previous information, as would be required to find an optimal strategy, then the graphical check simplifies considerably and we only need to check what is called *simple stability* (Dawid & Didelez, 2005).

Motivated by the question of identifying conditional (sequential) interventions, the identification of *conditional interventional distributions* has been considered (e.g. Pearl, 2000, section 4.2; Tian, 2004; Shpitser & Pearl, 2006) within the context of causal diagrams where all hidden variables are represented implicitly using bidirected edges. In particular, Shpitser & Pearl (2006) give necessary and sufficient criteria for this identification problem. Our approach is slightly different as we use influence diagrams, where the interventions are made explicit by suitable decision nodes and where unobservable variables are also explicitly shown by individual nodes (Dawid, 2002).

We do not consider in this paper the actual estimation and optimisation procedure required to find an optimal strategy, which is a numerically demanding task and a topic of its own. A regret–based approach to this has been proposed by Murphy (2003), while Robins (2004) suggests structural nested models (cf. also discussion by Moodie et al. (2007)). An application of the regret–based method to the anticoagulation problem described above can be found in Rosthoj et al. (2008).

The paper is organised as follows. In section 2 we introduce the notation used throughout. Optimal strategies are addressed in section 3. The general problem of identifying possibly conditional strategies is covered in section 4, where we present a simple sufficient criterion called simple stability as well as a more general criterion and explain how both can be checked graphically using influence diagrams. The latter reduces to simple stability when we consider optimal strategies as shown in section 5. We discuss and compare our approach in sections 6 and 7.

## 2 GENERAL SET–UP

Let $A_1, \ldots, A_N$ be variables that can be set by some intervention (action variables), while $L_1, \ldots, L_N$ (each of which can be vector valued) are additional observations / covariates, and let $L_{N+1} \equiv Y$ be the outcome of interest. To simplify exposition we restrict ourselves to the case of all variables being discrete. We assume that $\mathbf{L}^{\leq i} = (L_1, \ldots, L_i)$ can be observed before a decision about action $A_i$ is made. We will denote $\mathbf{L} = \mathbf{L}^{\leq N} = (L_1, \ldots, L_N)$ and similar for $\mathbf{A}$. A strategy $\mathbf{s} = (s_1, \ldots, s_N)$ consists of a set of functions that assign to any partial history $(\mathbf{a}^{<i}, \mathbf{l}^{\leq i}) = (a_1, \ldots, a_{i-1}, l_1, \ldots, l_i)$ a value $a_i = s_i(\mathbf{a}^{<i}, \mathbf{l}^{\leq i})$ in the space of $A_i$, and by following strategy $\mathbf{s}$ we mean that $A_i$ are set to $s_i(\mathbf{a}^{<i}, \mathbf{l}^{\leq i})$ by some intervention for all $i = 1, \ldots, N$. Let $\mathcal{S}$ be the set of relevant strategies (this set might be restricted e.g. due to feasibility). As mentioned above, a strategy where $s_i(\mathbf{a}^{<i}, \mathbf{l}^{\leq i}) \equiv s_i$ does not actually depend on $\mathbf{a}^{<i}, \mathbf{l}^{\leq i}$ is called unconditional, as the actual values that the $A_i$'s are set to do not depend on the observed history. Otherwise we call the strategy conditional. More generally we might also allow random (or stochastic) strategies, where $s_i(\mathbf{a}^{<i}, \mathbf{l}^{\leq i})$ specifies a distribution over the space of $A_i$ meaning that the intervention consists of drawing $a_i$ from this distribution and then setting $A_i = a_i$ by an intervention. Our framework allows for such randomised strategies (cf. Didelez et al. (2006) where this is used in a similar context, focusing on direct effects).

## 3 OPTIMAL STRATEGIES

If the distribution $p(y; \mathbf{s})$ of $Y$ when following strategy $\mathbf{s}$ is known, we can evaluate for any function $k(\cdot)$ the expectation $E\{k(Y); \mathbf{s}\}$; typically $k(\cdot)$ would be a loss function. This calculation can be implemented recursively. Define

$$f(\mathbf{a}^{\leq j}, \mathbf{l}^{\leq i}) = E\{k(Y) | \mathbf{a}^{\leq j}, \mathbf{l}^{\leq i}; \mathbf{s}\}$$

where $j = i$ or $j = i-1$ and $i = 1, \ldots, N+1$[1]. We have that $f(\mathbf{a}^{\leq N}, \mathbf{l}^{\leq N+1}) = k(y)$ and $f(\emptyset) = E\{k(Y); \mathbf{s}\}$, where starting with the former the latter can be obtained by iteratively applying the following operations for $i = N+1, \ldots, 1$

$$f(\mathbf{a}^{<i}, \mathbf{l}^{\leq i}) = \sum_{a_i} p(a_i | \mathbf{a}^{<i}, \mathbf{l}^{\leq i}; \mathbf{s}) \times f(\mathbf{a}^{\leq i}, \mathbf{l}^{\leq i}) \quad (1)$$

(note that $p(a_i | \mathbf{a}^{<i}, \mathbf{l}^{\leq i}; \mathbf{s})$ is by definition determined by the strategy $\mathbf{s}$ and equal to the indicator function $\mathbf{I}\{a_i = s_i(\mathbf{a}^{<i}, \mathbf{l}^{\leq i})\}$ unless $\mathbf{s}$ is a randomised strategy) and

$$f(\mathbf{a}^{<i}, \mathbf{l}^{<i}) = \sum_{l_i} p(l_i | \mathbf{a}^{<i}, \mathbf{l}^{<i}; \mathbf{s}) \times f(\mathbf{a}^{<i}, \mathbf{l}^{\leq i}). \quad (2)$$

This is exactly the procedure underlying the "extensive form" analysis of sequential decision theory (see e.g. Raiffa (1968)).

The optimal strategy $\mathbf{s}_{opt}$ is given by optimising $E\{k(Y); \mathbf{s}\} = f(\emptyset)$ over the set $\mathcal{S}$; assuming that $k(\cdot)$ is chosen such that large values are better, we have

$$\mathbf{s}_{opt} = \arg\max_{\mathcal{S}} E\{k(Y); \mathbf{s}\}.$$

The dynamic programming method to find an optimal strategy essentially proceeds as follows: starting with $i = N+1$ and working down to $i = 1$ find $a_i$ as function of $(\mathbf{a}^{<i}, \mathbf{l}^{\leq i})$ that maximises $f(\mathbf{a}^{\leq i}, \mathbf{l}^{\leq i})$. Hence, the optimal strategy will typically be conditional on past observations.

## 4 IDENTIFIABILITY

In practice we do not know the conditional distributions $p(l_i | \mathbf{a}^{<i}, \mathbf{l}^{<i}; \mathbf{s})$, $i = 1, \ldots, N+1$ for all $\mathbf{s} \in \mathcal{S}$ required to evaluate (2). Identifiability addresses the question whether data that have been gathered under an observational regime (which might be a sequentially randomised trial, or on observational study in the traditional meaning) can, in principle, inform us about these conditional distributions. We first address the question of identifiability of a single strategy **s**. It then follows that the optimal strategy can be identified if all strategies in $\mathcal{S}$ are identified so that the above optimisation can be carried out.

---

[1]If the subscript or superscript of a set is not defined then the set is defined to be $\emptyset$

In order to formalise the question of identification we introduce an indicator variable $\sigma$ for the regime, where $\sigma = o$ denotes the observational regime, under which the data is collected, and $\sigma = \mathbf{s}$ denotes the regime under which strategy **s** is followed. In a sequentially randomised study $p(a_i|\mathbf{a}^{<i}, \mathbf{l}^{\leq i}; o)$ would typically be known, whereas in a traditional observational study it would be unknown but might be estimable from data.

**Identifiability:** A strategy **s** is identified if we can obtain $E\{k(Y); \mathbf{s}\}$ uniquely from the joint distribution of $(\mathbf{A}, \mathbf{L}, Y)$ under $\sigma = o$.

Conditions for identifiability require a positivity condition, such that actions that arise from the strategy $a_i = s_i(\mathbf{a}^{<i}, \mathbf{l}^{\leq i})$ actually have a positive probability to occur under $\sigma = o$ if the history $(\mathbf{a}^{<i}, \mathbf{l}^{\leq i})$ has a positive probability to occur under $\sigma = \mathbf{s}$ (for a more formal definition of positivity see Dawid & Didelez (2005)); here, we will take positivity for granted.

### 4.1 SIMPLE STABILITY

It can easily be seen from (2) that a sufficient condition for identifiability of **s** is that the following conditional distributions are the same under both regimes

$$p(l_i|\mathbf{a}^{<i}, \mathbf{l}^{<i}; \mathbf{s}) = p(l_i|\mathbf{a}^{<i}, \mathbf{l}^{<i}; o),$$

$i = 1, \ldots, N+1$, whenever the conditioning event has positive probability under both regimes (Dawid & Didelez, 2005). This is called *simple stability* and also symbolised by

$$L_i \mathrel{\perp\!\!\!\perp} \sigma | (\mathbf{A}^{<i}, \mathbf{L}^{<i}), \qquad i = 1, \ldots, N+1 \quad (3)$$

where $\sigma$ takes values in $\{o, \mathbf{s}\}$. Simple stability is closely related to the *no unmeasured confounders* assumption (Robins, 1997). Note that the above notation for conditional independence (Dawid, 1979) has been generalised for problems involving decision variables by Dawid (2002), and can be checked graphically[2] on appropriate influence diagrams as shown in Figure 1, for example.

### 4.2 G–RECURSION

With simple stability, the target $E\{k(Y); \mathbf{s}\} = f(\emptyset)$ can now be obtained by iterating (1) and (2) starting

[2]using the *moralisation* criterion (cf. Cowell et al., 1999) or, equivalently, *d–separation* (Verma & Pearl, 1990)

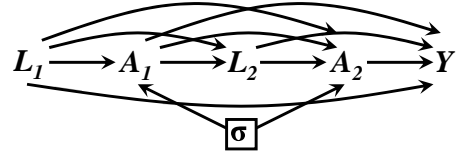

Figure 1: Example for simple stability.

with $i = N + 1$, where in (1) $p(a_i|\mathbf{a}^{<i}, \mathbf{l}^{\leq i}; \mathbf{s})$ is known by definition of the strategy **s** and (2) can be modified to

$$f(\mathbf{a}^{<i}, \mathbf{l}^{<i}) = \sum_{l_i} p(l_i|\mathbf{a}^{<i}, \mathbf{l}^{<i}; o) \times f(\mathbf{a}^{<i}, \mathbf{l}^{\leq i}).$$

due to simple stability. The name G–recursion has been coined by Robins (1986).

### 4.3 EXTENDED STABILITY

It will typically be difficult to believe in simple stability (3) without additional considerations. We might want to proceed by constructing a more complex but more acceptable model, typically including additional not necessarily observable variables, and investigate whether or not we can deduce the desired simple stability property. We denote the set of additional variables by **U**. *Extended stability* then holds if this set can be partitioned into $(U_1, \ldots, U_N)$ (each $U_i$ can be vector valued or the empty set) such that $\mathbf{U}^{<i}$ are not affected by an intervention in $A_i$, $i = 1, \ldots, N$, and

$$(L_i, U_i) \mathrel{\perp\!\!\!\perp} \sigma | (\mathbf{A}^{<i}, \mathbf{L}^{<i}, \mathbf{U}^{<i}), \quad i = 1, \ldots, N+1, \quad (4)$$

Clearly (4) implies that strategy **s** could be identified if **U** was observed as it is then just the same as simple stability w.r.t. $(\mathbf{A}, \mathbf{L}, \mathbf{U}, Y)$.

In many problems an extended stability assumption might be regarded as more reasonable and defensible than simple stability, so long as appropriate unobserved variables **U** are taken into account. For example, this might be the case if we believed that under $\sigma = o$ the actions $A_i$ were taken by a decision-maker who had access to variables in the set **U** as well as **L**.

Extended stability does not in general allow G–recursion if **U** is unobserved. However, it may do so if we can assume an (in)dependence structure on $(\mathbf{A}, \mathbf{L}, \mathbf{U}, Y)$ and $\sigma$ that allows us to deduce that simple stability (3) does hold. Dawid & Didelez (2005) give some further conditions for this and show how these can be verified graphically. As an example consider the graphs in Figure 2, where extended stability holds (here $L_1 = \emptyset$). In graph (a) simple stability is violated as $L_2 \mathrel{\not\perp\!\!\!\perp} \sigma | A_1$. In graph (b) the only change is that $U_1 = L_1$ can now be observed and then simple stability holds as $L_1 \perp\!\!\!\perp \sigma$, $L_2 \perp\!\!\!\perp \sigma | A_1, L_1$ and $Y \perp\!\!\!\perp \sigma | \mathbf{A}, \mathbf{L}$.

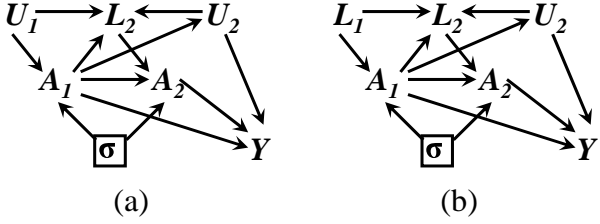

Figure 2: Examples for extended stability; in (a) simple stability is violated while in (b) it holds.

### 4.4 GENERAL CONDITIONS

A strategy **s** can be identified under weaker conditions than simple stability. This has been demonstrated by Pearl & Robins (1995) for the case of unconditional strategies. Next we extend the result to conditional (possibly stochastic) strategies.

We assume extended stability with respect to $(\mathbf{A}, \mathbf{L}, \mathbf{U}, Y)$ and define joint distributions $p_i(\cdot)$ as

$$p_i(\mathbf{A}, \mathbf{L}, \mathbf{U}, Y) = p(\mathbf{A}^{\leq i}, \mathbf{L}^{\leq i}, \mathbf{U}^{\leq i}; o) \\ \times p(\mathbf{A}^{>i}, \mathbf{L}^{>i}, \mathbf{U}^{>i}, Y | \mathbf{A}^{\leq i}, \mathbf{L}^{\leq i}, \mathbf{U}^{\leq i}; \mathbf{s})$$

Hence $p_0(\cdot) = p(\cdot; \mathbf{s})$ is the joint distribution under the strategy **s**, while $p_N(\cdot) = p(\cdot; o)$ is the joint distribution under the observational regime, where we exploit extended stability to obtain $p(Y | \mathbf{A}, \mathbf{L}, \mathbf{U}; \mathbf{s}) = p(Y | \mathbf{A}, \mathbf{L}, \mathbf{U}; o)$.

**Theorem 1.** Under extended stability, the strategy **s** is identified by G–recursion if

$$p_{i-1}(y | \mathbf{a}^{\leq i}, \mathbf{l}^{\leq i}) = p_i(y | \mathbf{a}^{\leq i}, \mathbf{l}^{\leq i}), \qquad i = 1, \ldots, N. \quad (5)$$

**Proof:** see appendix, and Dawid & Didelez (2005).

Property (5) can be paraphrased as the distribution of $Y$ given $\mathbf{a}^{\leq i}, \mathbf{l}^{\leq i}$ having to be the same regardless of whether $a_i$ has arisen out of the strategy **s** or from observation $o$, when we know that future actions will follow the strategy **s**.

A graphical check for (5) is more involved than for simple stability as it has to reflect the particular construction of the distributions $p_i$. This will be addressed in the next section.

### 4.5 GRAPHICAL CHECKS

If we express our subject matter background knowledge about the conditional independence structure among $(\mathbf{A}, \mathbf{L}, \mathbf{U}, Y)$ and the way we can intervene in $\mathbf{A}$ graphically we can check the above conditions for identifiability by simple graphical checks. Two approaches are possible. Firstly, we can augment the graph with the intervention indicator $\sigma$ as advocated in Pearl (1993), Lauritzen (2001), Dawid (2002); this augmented graph (influence diagram) will be denoted by $D$. Simple stability (3) wrt. observables $(\mathbf{A}, \mathbf{L}, Y)$ can, for example, be checked on such influence diagrams as in Figures 1 and 2. Secondly, and as is common in much of the mainstream causal literature, we can take the interventions in $\mathbf{A}$ as implicit and formulate graphical conditions involving $(\mathbf{A}, \mathbf{L}, \mathbf{U}, Y)$ only, omitting $\sigma$. We denote the graph which implicitly assumes extended stability with respect to sequential interventions in $\mathbf{A}$ by $D'$ (this is also called a causal graph with respect to $\mathbf{A}$ (Pearl, 2000). The graphs $D$ and $D'$ only differ in that the former has the additional decision node $\sigma$ with arrows into $\mathbf{A}$. It is easy to see that simple stability can therefore be checked on $D'$ by assessing whether $\mathbf{L}^{<i}$ satisfies the back–door criterion relative to $(\mathbf{A}^{<i}, L_i)$ (Pearl, 1995). This implies that the causal effects of each $A_i$ on later covariates $L_j$, $j > i$, are identified.

For the graphical check of (5) we first define the different parent sets for the actions $A_i$ under different regimes. Let $\text{pa}_o(A_i)$ be the parent nodes (excluding $\sigma$) of $A_i$ in $D$ when $\sigma = o$ and let $\text{pa}_\mathbf{s}(A_i)$ be the parent nodes (excluding $\sigma$) of $A_i$ in $D$ when $\sigma = \mathbf{s}$, i.e. if $s_i(\mathbf{a}^{<i}, \mathbf{l}^{\leq i})$ is constant in some of its arguments then these are not in $\text{pa}_\mathbf{s}(A_i)$. The two parent sets are not the same, as under $\sigma = o$ $A_i$ may depend on some variables in $\mathbf{U}^{\leq i}$, while under a strategy **s** we can obviously only take observable variables into account when choosing an action.

Now, we construct augmented graphs $D_i$ such that the only arrow out of the node $\sigma$ is into $A_i$, and such that the graph parents of the action variable $A_j$ are given by $\text{pa}_o(A_j)$ for $j < i$ and $\text{pa}_\mathbf{s}(A_j)$ for $j > i$ while the parent set for $A_i$ is the union of the parents under both regimes and $\sigma$. Such a graph represents the factorisation of the distribution $p_i$ constructed in section 4.4 if $A_i$ arises under $\sigma = o$ and of $p_{i-1}$ if $A_i$ is generated according to $\sigma = \mathbf{s}$. Let $[\cdot \perp\!\!\!\perp \cdot | \cdot]_{D_i}$ denote graph separation in $D_i$ then we have that (5) holds if

$$[Y \perp\!\!\!\perp \sigma | \mathbf{A}^{\leq i}, \mathbf{L}^{\leq i}]_{D_i} \quad (6)$$

This procedure is illustrated in Figure 3 for the example of graph (a) from Figure 2 assuming an unconditional intervention in $A_2$, i.e. $\text{pa}_\mathbf{s}(A_2) = \emptyset$. Hence there are no arrows into $A_2$ in $D_1$. We can easily see that $[Y \perp\!\!\!\perp \sigma | A_1]_{D_1}$ and $[Y \perp\!\!\!\perp \sigma | A_1, A_2, L_2]_{D_2}$ showing that a strategy where $A_2$ is chosen without taking

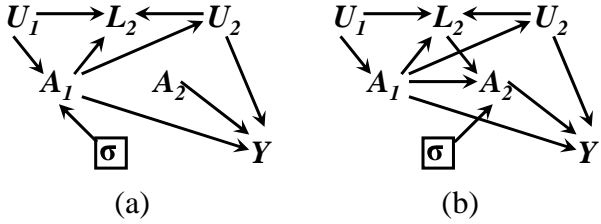

(a)                       (b)

Figure 3: Same example as in Figure 2(a); here (a) shows $D_1$ and (b) shows $D_2$ with uncond. intervention in $A_2$.

past covariates into account *is* identifiable even though simple stability as investigated in Figure 2 is violated.

In contrast, in the same example if we assume that the intervention $s_2$ in $A_2$ *does* depend on previous covariates, i.e. $\text{pa}_{\mathbf{s}(A_2)} = (A_1, L_2)$ then we have to modify $D_1$ as shown in Figure 4. Now we find that $[Y \perp\!\!\!\perp \sigma | A_1]_{D_1}$ is violated and we cannot guarantee that such a conditional strategy is identifiable.

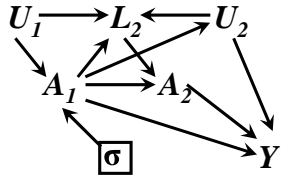

Figure 4: Same example as in Figure 3(a), $D_1$ with conditional intervention in $A_2$.

Pearl & Robins (1995) show that based on a causal diagram $D'$ (that does not include a node $\sigma$) and for unconditional sequential interventions sufficient graphical conditions are given by

$$[Y \perp\!\!\!\perp A_i | \mathbf{A}^{<i}, \mathbf{L}^{\leq i}]_{D'_i}, \tag{7}$$

where $D'_i$ is the graph $D'$ with all edges out of $A_i$ and all edges into $\mathbf{A}^{>i}$ removed. Comparing this with our procedure we can see that the idea is the same: deleting the edges out of $A_i$ corresponds to retaining only the back–door paths from $A_i$ as it is only these that are relevant when checking (6) due to $\sigma$ having only an arrow into $A_i$. Further, deleting every edge into $A_{i+1}, \ldots, A_N$ corresponds to changing the parent sets of these variables to only include the parents under $\mathbf{s}$. Note that if the interventions are unconditional then $A_j$ has no parents among $\mathbf{A}^{<j}, \mathbf{L}^{\leq j}, \mathbf{U}^{\leq j}$ under $\sigma = \mathbf{s}$.

Hence, an immediate extension of Pearl & Robins' approach, that has also been suggested by Robins (1997), to the case of conditional interventions is given by modifying the meaning of $D'_i$ in (7) so that only edges into $A_j$, $j = i+1, \ldots, N$, are deleted that the intervention $s_j$ does not depend on. This will always be the case for all edges from $\mathbf{U}^{\leq j}$ into $A_j$ because the intervention can obviously not be a function of unobserved quantities. This is illustrated in Figure 5 for the same example as above with a conditional intervention at $A_2$. We can see that $[Y \perp\!\!\!\perp A_1]_{D'_1}$ is violated while $[Y \perp\!\!\!\perp A_2 | A_1, L_2]_{D'_2}$ holds.

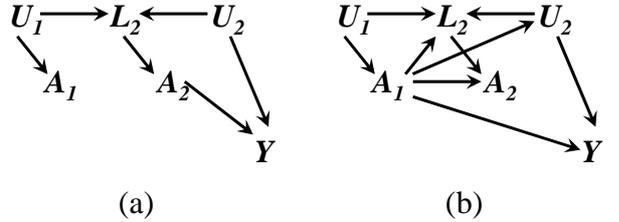

(a)                       (b)

Figure 5: Pearl & Robins' check when $s_2$ is conditional, (a) $D'_1$ and (b) $D'_2$

## 5 IDENTIFIABILITY OF OPTIMAL STRATEGIES

As argued in section 3, the optimal strategy will typically be a conditional strategy, i.e. a strategy that must allow action $a_i = s_i(\mathbf{a}^{<i}, \mathbf{l}^{\leq i})$ to depend on all of $(\mathbf{a}^{<i}, \mathbf{l}^{\leq i})$ (notice that dependence on $\mathbf{a}^{<i}$ is only relevant for random strategies as otherwise $a_j$ is a function of $\mathbf{l}^{\leq j}$, $j = 1, \ldots, i-1$ anyway). We show now that in this case the more general conditions of section 4.4 reduce to simple stability.

First we need some regularity assumptions.

**Assumption 1.** We assume $\text{pa}_{\mathbf{s}}(A_i) \subset \text{pa}_o(A_i)$ for all $i = 1, \ldots, N$.

The assumption means that the parents of $A_i$ when we follow the strategy $\mathbf{s}$ are a subset of its parents under the observational regime. This can easily be satisfied by redefining, if necessary, $\text{pa}_o(A_i)$ as $\text{pa}_o(A_i) \cup \text{pa}_s(A_i)$, with any added parents having no effect on the conditional probabilities for $A_i$ under $o$. Note that adding edges from *observable* nodes into action nodes cannot destroy identifiability.

**Assumption 2.** Each $L_1, \ldots, L_N$ is an ancestor of $Y$ in the graph $D_0$, where the parents of $A_i$ are as under strategy $\mathbf{s}$, $i = 1, \ldots, N$.
This means that the covariates predict $Y$ when we follow the strategy $\mathbf{s}$.

**Remark.** Assumption 2 is implied if
(i) each $A_1, \ldots, A_N$ is an ancestor of $Y$ in $D_0$, *and*

(ii) each $L_1, \ldots, L_N$ is an ancestor of some $A_j$ in $D_0$.

As we want to investigate whether the actions affect $Y$, part (i) will be plausible because otherwise we have at least one action of which we know *a priori* that it does not predict $Y$ and this would not typically be included in the investigation. Part (ii) is relevant in the context of optimal strategies as these must in principle be allowed to depend on all previous observable covariates.

**Theorem 2.** Suppose Assumptions 1 and 2 hold. Then if the graphical check of (5) succeeds, the problem exibits simple stability with respect to $(\mathbf{A}, \mathbf{L}, Y)$. **Proof:** see appendix.

In consequence we can say that if the target is to find an optimal strategy and we are asking whether this is possible from a given data situation we do not need to apply the more complex graphical check of (5) but can just check simple stability, which has the advantage that it can be seen from a single graph like e.g. Figure 2(b). This also implies that a sufficient criterion for identification of an optimal strategy is that the causal effects of each $A_i$ on later covariates $L_j$, $j > i$, can be identified by the back–door criterion as mentioned in section 4.5.

## 6 RELATION TO OTHER APPROACHES

It has been argued that conditional strategies can be identified if *conditional intervention distributions* can be identified (Pearl, 2000, section 4.2). For the case of a non–stochastic conditional strategy $\mathbf{s}$ that fixes $\mathbf{a} = \mathbf{s}(\mathbf{l})$ this is seen as follows. We have that

$$p(y; \mathbf{s}) = \sum_{\mathbf{l}} p(y|\mathbf{l}; \mathbf{s}) p(\mathbf{l}; \mathbf{s}) \quad (8)$$

(the recursive version of which is based on (2)). As $\mathbf{l}$ in $p(y|\mathbf{l}; \mathbf{s})$ is given, we have $p(y|\mathbf{l}; \mathbf{s}) = p(y|\mathbf{l}; \mathbf{a})$, where $\sigma = \mathbf{a}$ denotes an unconditional strategy and $\mathbf{a} = \mathbf{s}(\mathbf{l})$. Hence we can identify $p(y; \mathbf{s})$ if we can identify $p(y|\mathbf{l}; \mathbf{a})$ for every sequence $\mathbf{l}$ and every unconditional strategy $\sigma = \mathbf{a}$. Shpitser & Pearl (2006) give a sound and complete algorithm to identify such conditional intervention distributions, like $p(y|\mathbf{l}; \mathbf{a})$, which outputs FAIL iff the problem is not identified. This takes a semi–Markovian graph as input which is based on a causal model.

However, for the whole $p(y; \mathbf{s})$ to be identifiable by (8) we also need $p(\mathbf{l}; \mathbf{s})$ to be identifiable, where it is important to note that due to the sequential nature of the problem some covariates will be affected by earlier actions and hence we cannot assume $p(\mathbf{l}; \mathbf{s}) = p(\mathbf{l}; o)$. Tian (2004) gives an example where $p(y|\mathbf{l}; \mathbf{s})$ is identified while $p(\mathbf{l}; \mathbf{s})$ is not. Hence, identifiability of conditional sequential plans is not covered by the identifiability of conditional interventional distributions alone.

Our results do not provide a necessary criterion for identifiability. However, they do not require a semi–Markovian graph nor a causal model (which in our case would assume that we can intervene in any of $L_1, \ldots, L_N$ as well as in $\mathbf{A}$) as long as the background knowledge can be encoded in a DAG on $(\mathbf{A}, \mathbf{L}, \mathbf{U}, Y, \sigma)$. More importantly, as mentioned earlier, simple stability implies that the effect of each action on later covariates $p(\mathbf{l}; \mathbf{s})$ as well as the conditional intervention distribution $p(y|\mathbf{l}; \mathbf{s})$ are identifiable, so that (8) applies.

## 7 DISCUSSION

We have addressed the question of identifying optimal sequential strategies within the framework of decision theory. Simple stability (3) provides a straightforward graphical check for identifiability on a single influence diagram, and the more involved check for the conditions in section 4.4 that extends Pearl & Robins (1995) approach is in fact not more general.

We would like to point out that even though the target of inference $E(k(Y); \mathbf{s})$ can be constructed using the G–recursion, it is in practice not advisable to estimate an optimal strategy (when it *is* identified) by estimating the individual factors of the G–recursion formula (Robins & Wasserman, 1997). This has motivated the alternative approaches such as suggested by Murphy (2003) and Robins (2004). The reasoning regarding identifiability, however, remains valid.

## APPENDIX

**PROOF OF THEOREM 1**

This is a special case of a more general result (Dawid and Didelez, 2005, § 7.1). The following argument is specialised to the current context, and assumes that events conditioned on have positive probability.

Similar to section 3, define

$$\begin{aligned} f(\mathbf{a}^{\leq i}, \mathbf{l}^{\leq i}) &= E_i\{k(Y) \mid \mathbf{a}^{\leq i}, \mathbf{l}^{\leq i}\} \\ f(\mathbf{a}^{< i}, \mathbf{l}^{\leq i}) &= E_{i-1}\{k(Y) \mid \mathbf{a}^{< i}, \mathbf{l}^{\leq i}\}, \end{aligned}$$

where $E_i$ denotes expectation under distribution $p_i$. In particular, $f(\emptyset) = E\{k(Y)|\mathbf{s}\}$, which is what we wish to compute; while $f(\mathbf{a}^{\leq N}, \mathbf{l}^{\leq N}) = E\{k(Y)|\mathbf{a}^{\leq N}, \mathbf{l}^{\leq N}; o\}$ is estimable from the observational data.

By extended stability, the distribution of $(L_i, U_i)$ given $(\mathbf{A}^{<i}, \mathbf{L}^{<i}, \mathbf{U}^{<i})$ is the same under both the observational regime $o$ and the strategy $\mathbf{s}$. It then follows from the definition of $p_{i-1}$ that the joint distribution of $(\mathbf{A}^{<i}, \mathbf{L}^{\leq i}, \mathbf{U}^{\leq i})$ is the same under $p_{i-1}$ as in $o$, whence in particular $p_{i-1}(l_i|\mathbf{a}^{<i}, \mathbf{l}^{<i}) = p(l_i|\mathbf{a}^{<i}, \mathbf{l}^{<i}; o)$. Thus, $f(\mathbf{a}^{<i}, \mathbf{l}^{<i}) =$

$$\begin{aligned} &\sum_{l_i} p_{i-1}(l_i|\mathbf{a}^{<i}, \mathbf{l}^{<i}) \times E_{i-1}\{k(Y) \mid \mathbf{a}^{<i}, \mathbf{l}^{\leq i}\} \\ =\ &\sum_{l_i} p(l_i|\mathbf{a}^{<i}, \mathbf{l}^{<i}; o) \times f(\mathbf{a}^{<i}, \mathbf{l}^{\leq i}). \qquad (9) \end{aligned}$$

Also, from the construction of $p_{i-1}$, the fact that $p(a_i|\mathbf{a}^{<i}, \mathbf{l}^{\leq i}; \mathbf{s})$ is fully determined independently of $\mathbf{u}^{\leq i}$, and (5), $f(\mathbf{a}^{<i}, \mathbf{l}^{\leq i}) =$

$$\sum_{a_i} p_{i-1}(a_i \mid \mathbf{a}^{<i}, \mathbf{l}^{\leq i}) \times E_{i-1}\{k(Y) \mid \mathbf{a}^{\leq i}, \mathbf{l}^{\leq i}\}$$

$$= \sum_{a_i} p(a_i \mid \mathbf{a}^{<i}, \mathbf{l}^{\leq i}; \mathbf{s}) \times f(\mathbf{a}^{\leq i}, \mathbf{l}^{\leq i}), \qquad (10)$$

where $E_{i-1}\{k(Y)|\mathbf{a}^{\leq i}, \mathbf{l}^{\leq i}\} = f(\mathbf{a}^{\leq i}, \mathbf{l}^{\leq i})$ due to (5). Together, (9) and (10) show that $f$ can be computed by $G$-recursion: starting with $f(\mathbf{a}^{\leq N}, \mathbf{l}^{\leq N}) = E\{k(Y) \mid \mathbf{a}^{\leq N}, \mathbf{l}^{\leq N}; o\}$, we will exit the recursion with $f(\emptyset) = E\{k(Y) \mid \mathbf{s}\}$.

## PROOF OF THEOREM 2

We introduce the following abbreviations for certain graph-separation properties:

$$\begin{aligned}
\gamma_i &: [Y \perp\!\!\!\perp \sigma | (\mathbf{L}^{\leq i}, \mathbf{A}^{\leq i})]_{D_i} \\
\alpha_{ij} &: [L_{j+1} \perp\!\!\!\perp_i \sigma | (\mathbf{L}^{\leq i}, \mathbf{A}^{\leq i})]_{D_i} \\
\alpha_k &: \alpha_{ij} \text{ holds for all } 0 \leq i \leq j \leq k \\
\beta_k &: [L_{k+1} \perp\!\!\!\perp \sigma | (\mathbf{A}^{\leq k}, \mathbf{L}^{\leq k})]_D \\
H_k &: \alpha_k \Rightarrow \beta_k
\end{aligned}$$

Also, we use $\text{man}_i(A, B, C)$ to denote the moralised subgraph of $D_i$ on the set $\text{An}(A \cup B \cup C)$[3].

**Lemma 1.** Suppose Assumptions 1 and 2 hold. Then for $1 \leq i \leq j \leq N$, $\gamma_i \Rightarrow \alpha_{ij}$.

**Proof:** Fix $i$, and suppose that for some $j \geq i$, $\alpha_{ij}$ fails. Then there is a path $\pi_1$ in $\text{man}_i(\mathbf{A}^{\leq i}, \mathbf{L}^{\leq i}, L_{j+1})$ joining $\sigma$ to some $L \in L_{j+1}$ avoiding $(\mathbf{A}^{\leq i}, \mathbf{L}^{\leq i})$. Since by Assumption 2 we have $L_{j+1}$ is an ancestor of $Y$ in $D_i$, $\pi_1$ is a path in $\text{man}_i(\mathbf{A}^{\leq i}, \mathbf{L}^{\leq i}, Y)$ with the same property. But again by Assumption 2, there is a descending path $\pi_2$ in $D_i$ joining $L$ to $Y$ avoiding $(\mathbf{A}^{\leq i}, \mathbf{L}^{\leq i})$ since these are non–descendants of $L_{j+1}$. Hence, the concatenation of $\pi_1$ and $\pi_2$ is a path in $\text{man}_i(\mathbf{A}^{\leq i}, \mathbf{L}^{\leq i}, Y)$ joining $\sigma$ and $Y$ while avoiding $(\mathbf{A}^{\leq i}, \mathbf{L}^{\leq i})$ showing that $\gamma_i$ fails.

**Lemma 2.** Suppose Assumption 1 holds. Then $H_k$ holds, for all $0 \leq k \leq N$.

**Proof:** The proof is by induction on $k$. $H_0$ holds since $\beta_0$ holds by extended stability. Now fix $k \leq N$ and suppose that $H_l$ holds for all $0 \leq l < k$ and argue that $H_k$ holds. To do so we assume that $\alpha_k$ holds, and argue that $\beta_k$ follows.

Since for all $0 \leq l < k$ $\alpha_l$ must hold, from $H_l$ so must $\beta_l$.

Before proceeding we introduce some notation. For $1 \leq j \leq k$, we denote by $M_j$ (resp. $N_j$) the moralised ancestral graph of $(\mathbf{A}^{\leq j}, \mathbf{L}^{\leq j}, L_{k+1})$ in $D$ (resp. in $D_j$). We write $V_1 \searrow V_2$ to denote that there is a descending directed path in $D$ from node $V_1$ to node $V_2$ whose intermediate nodes are all in $\mathbf{U}$.

Now suppose $\beta_k$ fails, then there exists a path $\pi$ in $M_k$ having the following property:
**P**: $\pi$ connects $\sigma$ to $L_{k+1}$, all intermediate nodes being in $\mathbf{U}$.

Let $\Pi_j$ denote the property that any moral link in $\pi$ is due to a common child in $C$ in $(\mathbf{A}^{\leq j}, \mathbf{L}^{\leq j})$.

If $\pi$ contains a moral link due to a common child $C \notin (\mathbf{A}^{\leq k}, \mathbf{L}^{\leq k})$ then $C \in \mathbf{U}$, and we can modify $\pi$ by adding the intermediate node $C$ and removing the initial moral link, while still satisfying property **P**. Proceeding in this way for all such moral links we see that we can suppose that $\Pi_k$ holds.

Let $U_1 \in \mathbf{U}$ be the first internal node in $\pi$ after $\sigma$: this must exist because $L_{k+1}$ and $\sigma$ are not directly connected, nor can they have a common child in $M_k$. The link $\sigma$—$U_1$ can only be a moral link due to a common child $A_i$ for some $i \leq k$. Then we cannot have $U_1 \searrow L_{k+1}$ since this would create a path violating $\alpha_{ik}$.

Let $U_2$ be the first internal node, if one exists, in $\pi$ that is not an ancestor of $(\mathbf{A}^{\leq k}, \mathbf{L}^{\leq k})$ in $D$. Then we must have $U_2 \searrow L_{k+1}$ by some path $\pi_2$, say. In particular $U_1 \neq U_2$. Denote by $\pi_1$ the section of $\pi$ between $\sigma$ and $U_2$, and let $\pi_1^o$ be this section stripped of its endpoints. Since $U_1 \in \pi_1^o$, it is not empty. Now replace the original section section of $\pi$ from $U_2$ to $L_{k+1}$ by $\pi_2$. The new path will still possess properties **P** and $\Pi_k$. We replace $\pi$ by the concatenation of $\pi_1$ and $\pi_2$ which will be renamed $\pi$ for the sequel.

Consider now the hypotheses $(Q_j)$, $0 \leq j \leq k$, defined by $Q_j : \pi_1^o \subset \text{an}(\mathbf{A}^{\leq j}, \mathbf{L}^{\leq j})$ and property $\Pi_j$ holds. We shall show that $Q_j \Rightarrow Q_{j-1}$.

Thus suppose $Q_j$. There cannot exist $U \in \pi_1^o$ with $U \searrow A_j$ since this would violate $\alpha_{jk}$. We deduce that $\pi_1^o \subset \text{an}(\mathbf{A}^{\leq j}, \mathbf{L}^{\leq j})$ and any moral link in $\pi$ must be due to a common child in $(\mathbf{A}^{<j}, \mathbf{L}^{\leq j})$.

Now $Q_j$ implies that $\pi$ is a path in $M_j$. Then there cannot exist $U \in \pi_1^o$ with $U \searrow L_j$ since this would violate $\beta_{j-1}$. We deduce $Q_{j-1}$.

Now $Q_k$ holds by construction. Applying the above repeatedly we deduce $Q_0$, which can only hold if $\pi_1^o$ is empty. But $U_1 \in \pi_1^o$. This contradiction proves that $\beta_k$ holds and so we have demonstrated $H_k$. The desired result follows by induction.

**Proof of Theorem 2:** The graphical check succeeds just when we can show $\gamma_i$, $1 \leq i \leq N$. By the above Lemmas, $\alpha_N$ must hold which implies that $\beta_j$, $1 \leq j \leq N$ does. This is exactly the condition for simple stability.

---

[3]We use the moralisation criterion here to verify graph separation in DAGs